\documentclass[sn-mathphys,Numbered]{sn-jnl}

\usepackage{graphicx}%
\usepackage{multirow}%
\usepackage{amsmath,amssymb,amsfonts}%
\usepackage{amsthm}%
\usepackage{mathrsfs}%
\usepackage[title]{appendix}%
\usepackage{xcolor}%
\usepackage{textcomp}%
\usepackage{manyfoot}%
\usepackage{booktabs}%
\usepackage{algorithm}%
\usepackage{algorithmicx}%
\usepackage{algpseudocode}%
\usepackage{listings}%
\usepackage{subcaption}
\usepackage{caption}
\usepackage{adjustbox}
\usepackage{tabularx}
\usepackage{array}
\usepackage{mathtools}
\usepackage{lineno} 
\usepackage{float}

\theoremstyle{thmstyleone}%
\theoremstyle{thmstyletwo}%

\theoremstyle{thmstylethree}%

\definecolor{gold}{RGB}{255, 215, 0}
\definecolor{silver}{RGB}{192, 192, 192}
\definecolor{copper}{RGB}{184, 115, 51}
\definecolor{b1}{RGB}{186, 216, 242}
\definecolor{b2}{RGB}{232, 242, 251}
\definecolor{r1}{RGB}{250, 228, 227}

\begin{document}


\title[Article Title]{Academically intelligent LLMs are not necessarily socially intelligent}


\author[1,2]{\fnm{Ruoxi} \sur{Xu}}\email{ruoxi2021@iscas.ac.cn}
\author[2]{\fnm{Hongyu} \sur{Lin}}\email{hongyu@iscas.ac.cn}
\author[2,3]{\fnm{Xianpei} \sur{Han}}\email{xianpei@iscas.ac.cn}
\author[2,3]{\fnm{Le} \sur{Sun}}\email{sunle@iscas.ac.cn}
\author[1]{\fnm{Yingfei} \sur{Sun}}\email{yfsun@ucas.ac.cn}

\affil[1]{\orgname{School of Electronic, Electrical and Communication Engineering, University of Chinese Academy of Sciences}, \orgaddress{\city{Beijing}, \country{China}}}
\affil[2]{\orgname{Chinese Information Processing Laboratory, Institute of Software, Chinese Academy of Sciences}, \orgaddress{\city{Beijing}, \country{China}}}
\affil[3]{\orgname{State Key Laboratory of Computer Science, Institute of Software, Chinese Academy of Sciences}, \orgaddress{\city{Beijing}, \country{China}}}


\abstract{
The academic intelligence of large language models (LLMs) has made remarkable progress in recent times, but their social intelligence performance remains unclear.
Inspired by established human social intelligence frameworks, particularly Daniel Goleman's social intelligence theory, we have developed a standardized social intelligence test based on real-world social scenarios to comprehensively assess the social intelligence of LLMs, termed as the Situational Evaluation of Social Intelligence (SESI).
We conducted an extensive evaluation with 13 recent popular and state-of-art LLM agents on SESI.
The results indicate the social intelligence of LLMs still has significant room for improvement, with superficially friendliness as a primary reason for errors. Moreover, there exists a relatively low correlation between the social intelligence and academic intelligence exhibited by LLMs, suggesting that social intelligence is distinct from academic intelligence for LLMs. Additionally, while it is observed that LLMs can't ``understand'' what social intelligence is, their social intelligence, similar to that of humans, is influenced by social factors.}


\maketitle

\section{Introduction}
\label{intro}

The ability to understand and manage social relationships is one fundamental dimension of human intelligence, commonly denoted as social intelligence~\citep{thorndike1920}. Social intelligence enables humans to reduce conflicts and foster cooperation, thus navigating the social world. It not only correlates closely with individual success and life satisfaction~\citep{joseph2010social, zakirova2014success}, but also is one of the most important ingredients in humans' survival as a species in the long run~\citep{albrecht2006social}.

As a core component of human intelligence, social intelligence stands as an indispensable milestone on the path to achieving artificial general intelligence (AGI)~\citep{sterelny2007social}.
On one hand, social intelligence is necessary for effective interaction between intelligent agents and humans~\citep{dautenhahn1995getting}, with its significance becoming increasingly pronounced as AI technology continues to advance~\citep{zhao2023survey} and intelligent agents find increasing applications in our daily lives.
For example, the envisioned scenarios for intelligent systems in the "real world," such as welfare robots, household robots, and robots collaborating to solve common problems, heavily rely on effective communication and collaboration among artifacts as well as between artifacts and humans. This is particularly evident in cases where intelligent systems are expected to support humans in tasks involving numerous social interactions, such as serving as home tutors.
On the other hand, social intelligence provides the foundation for artificial intelligence systems, particularly Large Language Models (LLMs), to deeply learn, as language is inherently social, and meaning is constructed through social interactions~\citep{wittgenstein2019philosophical}.
Moreover, social intelligence is closely associated with crucial issues of AI alignment and governance.
Individuals with high social intelligence can effectively manage conflicts between individual and group objectives~\citep{korinek2022aligned}, which is precisely the essence of most social alignment issues. Individuals with high social intelligence can also avoid toxic behaviors that make others feel diminished, incompetent, intimidated, angry, frustrated, or guilty, by equipping awareness of the impact on others~\citep{albrecht2006social}.
Therefore, by strengthening research on social intelligence, we can better guide artificial intelligence towards a more intelligent and social direction, realizing a future of mutually beneficial human-machine collaboration.

While the importance of social intelligence is widely acknowledged~\citep{hovy2021importance}, evaluating it within recently developed advanced AI systems, particularly large language models such as ChatGPT~\citep{openai2021chatgpt, openai2023gpt}, Claude~\citep{anthropic2023claude}, and LLaMA~\citep{llama, llama2}, remains limited.
The current research predominantly focuses on the investigation of academic intelligence in LLMs, showcasing their high performance in social isolated tasks, such as logic, automated theorem proving, diagnostic reasoning and so on~\citep{chang2023survey, sarkisyan2023evaluation}. In contrast, the social intelligence of LLMs, crucial for real-world applications, is often perceived as a "side effect" and has not been comprehensively established in a robust manner. 
Some researchers assess the social intelligence of LLMs based on classic tests of human social intelligence, such as ToMi~\citep{le2019revisiting}, which tests whether models can distinguish their own and others' cognitive states in scenarios of information asymmetry, and FauxPas~\citep{shapira2023well}, which tests whether models can provide correct responses to questions involving faux pas situations. These well-established tests have a long history, making it likely that LLMs have been exposed to and trained on them, raising challenges in discerning whether models truly possess a generalizable understanding of social factors~\citep{shapira2023clever}.
Some other researchers assesses social intelligence of LLMs in the context of social factor understanding, exemplified by datasets such as SocialIQA~\citep{sap2019social}, SocKET~\citep{choi2023llms} and SECEU~\citep{wang2023emotional}. These datasets focus on assessment of social awareness, the ability to comprehend and track agents' inner states, such as emotions, beliefs, motivations and so on, while ignoring social facility, the ability to act smoothly and efficiently in relationships, which is necessary to guarantee fruitful interactions.
There are also two innovative benchmarks, SOTOPIA~\citep{zhou2023sotopia} and EmoBench~\citep{sabour2024emobench}, involving the application of social factors. However, they either employ manually crafted social contexts and goals, introducing subtle differences from real-world interactive scenarios, or solely focus on a single social factor, thereby limiting the ability to comprehensively assess social intelligence.
Therefore, there is a need for a dynamic and comprehensive benchmark to go beyond existing benchmarks, in order to fully assess the social intelligence of LLMs.

\begin{figure}[t]
  \centering
  \includegraphics[width=1\textwidth]{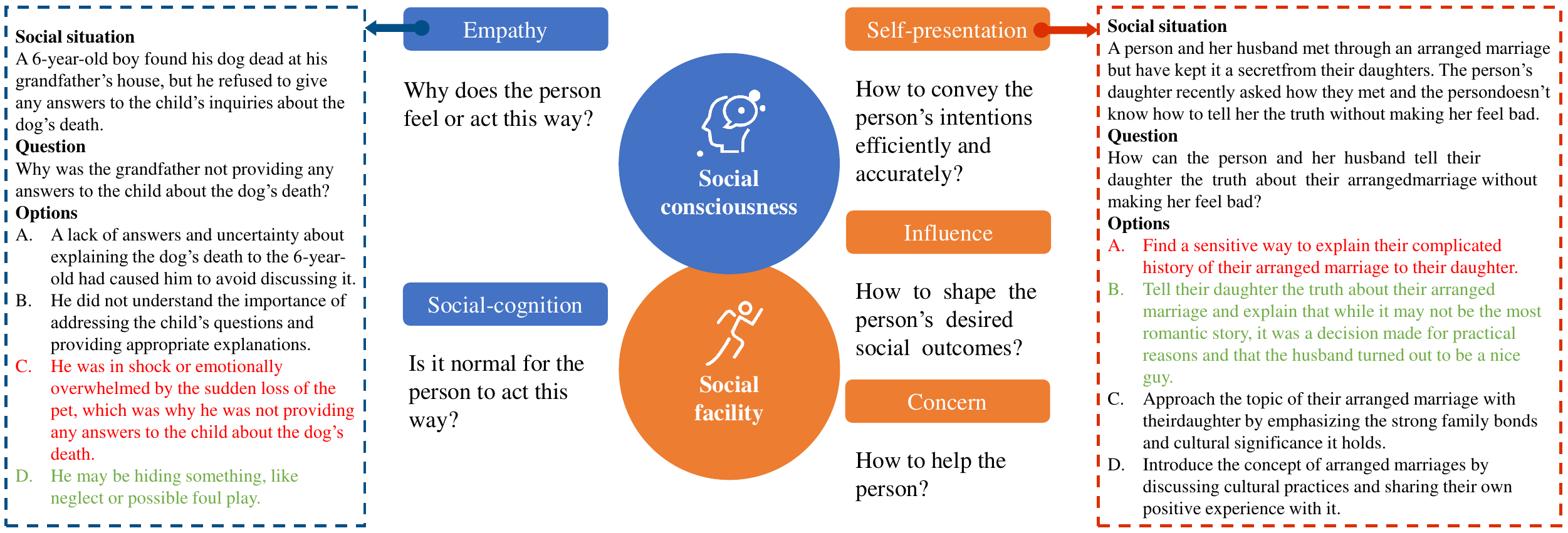}
  \caption{Overview of the situational evaluation of social intelligence. \textcolor{green}{Green} indicates the correct answer and \textcolor{red}{red} indicates the wrong answer selected by gpt-3.5-turbo-0613 model.}
  \label{fig:intro}
\end{figure}

To fill the gap, we first propose a social intelligence framework to comprehensively describe and evaluate the social intelligence of LLMs, which is inspired by established human social intelligence frameworks, including the S.P.A.C.E theory~\citep{albrecht2006social} and Daniel Goleman's social intelligence theory~\citep{daniel2006social}. The social intelligence framework posits that the social intelligence of LLMs comprises two categories: social awareness and social facility, further delineated into five subcategories, as shown in Figure~\ref{fig:intro}.
Following this, we developed the Situational Evaluation of Social Intelligence (SESI), which serves as a comprehensive, challenging benchmark for assessing the social intelligence of LLMs in real and complex social situations. For each sub-capacity of social intelligence mentioned earlier, the benchmark offers 100 corresponding questions as test items. Specifically, the social scenarios in the benchmark are derived from authentic requests for assistance posted by users on Reddit, with the correct answers determined based on the consensus of the top five most endorsed responses. This approach ensures the incorporation of genuine, complex social situations and allows for flexible and diverse solutions. 
Compared to the previously mentioned benchmarks, SESI possesses two distinctive advantages: 1) comprehensive, as our benchmark is grounded in established social intelligence theoretical framework, comprehensively assessing all the abilities encompassed by social intelligence; 2) dynamic, as test questions in our benchmark can be automatically generated based on Reddit Q\&A posts. This allows for automatic updates over time, representing a core distinction from previous evaluations conducted on static datasets.

We then conducted an evaluation of a spectrum of mainstream and widely-adopted LLMs on SESI, and obtained the following findings: 
1) The social intelligence of LLMs still has significant room for improvement, as evidenced by the best-performing model, gpt-3.5-turbo-0613, which achieves only 55.2\% performance. 
2) The social intelligence of LLMs is distinct from academic intelligence, warranting investigation as a separate form of intelligence. 
3) LLMs are superficially friendly, following fixed friendly patterns without grounding them in real social situations, which is the main reason for the errors made by LLMs in social judgments.
4) LLMs can't ``understand'' what social intelligence means, as evidenced by their inconsistent social intelligence level with prompts.
5) Social intelligence of LLMs, similar to that of human beings, is influenced by social factors, including personality, gender, social role and person.

\section{Results}

\begin{table}[t] \footnotesize
\setlength{\tabcolsep}{0pt} 
\begin{tabular}{c|c|cc|cc|cc|cc|c|c}
\toprule
\multirow{2}{*}{Series} & \multirow{2}{*}{Model} & \multicolumn{2}{c|}{Knowledge} & \multicolumn{2}{c|}{Reasoning} & \multicolumn{2}{c|}{Comprehension} & \multicolumn{2}{c|}{Math} & Safety & SI \\
 & & NQ & MMLU & BBH & WinoGrande & RACE-h & DROP & GSM8K & MATH & TruthfulQA & SESI \\ \midrule
\multirow{6}{*}{GPT} & gpt-4-0613 & \colorbox{b2}{48.6} & \colorbox{b1}{81.3} & \colorbox{b1}{84.6} & \colorbox{b1}{87.1} & \colorbox{b1}{91.8} & \colorbox{b1}{87.4} & \colorbox{b1}{92.1} & \colorbox{b1}{34.9} & \colorbox{b1}{79.1} & \colorbox{b2}{54.4} \\
 & gpt-3.5-turbo-0613 & 38.8 & \colorbox{b2}{67.4} & 68.1 & 55.3 & 81.2 & 53.7 & \colorbox{b2}{76.3} & 15 & \colorbox{b2}{61.4}
 & \colorbox{b1}{55.2} \\
 & text-davinci-003  & 38.1 & 63.7 & \colorbox{b2}{69} & \colorbox{b2}{70.6} & 79.5 & 56.3 & 59.4 & 15.6 & 52.2 & 38 \\
 & text-davinci-002& 28.2 & 62.1 & 66 & 65.5 & 80.5 & 47.5 & 47.3 & 8.5 & 47.8 & 42.8 \\
 & text-davinci-001  & 23.5 & 46.7 & 38.6 & 54.6 & 44.3 & 33.1 & 15.6 & \colorbox{r1}{0} & 54.2 & 36.9 \\
 & davinci & \colorbox{r1}{17.8} & 34.3 & 39.1 & 48 & 35 & \colorbox{r1}{16.5} & 12.1 & \colorbox{r1}{0} & 21.4 & \colorbox{r1}{0.4} \\
\midrule
\multirow{3}{*}{LLaMA2} & llama-2-70b-chat & 40.5 & 42.5 & 55.1 & 58.5 & 77 & \colorbox{b2}{58.7} & 56.9 & 6 & 38.3 & 49.4 \\
 & llama-2-13b-chat & 35.5 & 28.5 & 34.6 & 48.5 & 71.3 & 56.3 & 23.1 & 3.5 & 40.7 & 39.2 \\
 & llama-2-7b-chat  & 28 & 26.4 & \colorbox{r1}{30.1} & 46.5 & 55.7 & 45.3 & \colorbox{r1}{6.1} & 0.5 & \colorbox{r1}{16} & 41.6 \\
\midrule
\multirow{2}{*}{Vicuna} & vicuna-33b  & 33 & \colorbox{r1}{24.7} & 48.1 & 44.5 & \colorbox{r1}{29.3} & 55.2 & 47.7 & 1.5 & 30.9 & 32.4 \\
 & vicuna-13b & 24.5 & 45.4 & 57.4 & \colorbox{r1}{38.5} & 44.3 & 43 & 41.5 & 3 & 32.1 & 37.6 \\
\midrule
\multirow{2}{*}{Mistral} & mixtral-8x7b-instruct & \colorbox{b1}{49.5} & 57.1 & 59.3 & 57.5 & \colorbox{b2}{82.2} & 51.5 & 67.7 & \colorbox{b2}{23.5} & 56.8 & 50.8 \\
 & mixtral-7b-instruct & 21.5 & 46 & 49 & 46 & 62.6 & 40.8 & 41.5 & 5 & 48.1 & 39.5 
\\ \bottomrule
\end{tabular}
\caption{Evaluation results on representative academic intelligence benchmarks and SESI benchmark. The \colorbox{b1}{blue} represents the \colorbox{b1}{best-performing} models on the same benchmark, the \colorbox{b2}{light blue} represents the \colorbox{b2}{second-best-performing} models and the \colorbox{r1}{red} indicates the \colorbox{r1}{worstperforming} models. As indicated in the table, the academic intelligence of LLM agents can not accurately or at least comprehensively mirror their competence in social intelligence.}
\label{tab:si}
\end{table}

\begin{figure}[t]
  \centering
  \includegraphics[width=0.5\textwidth]{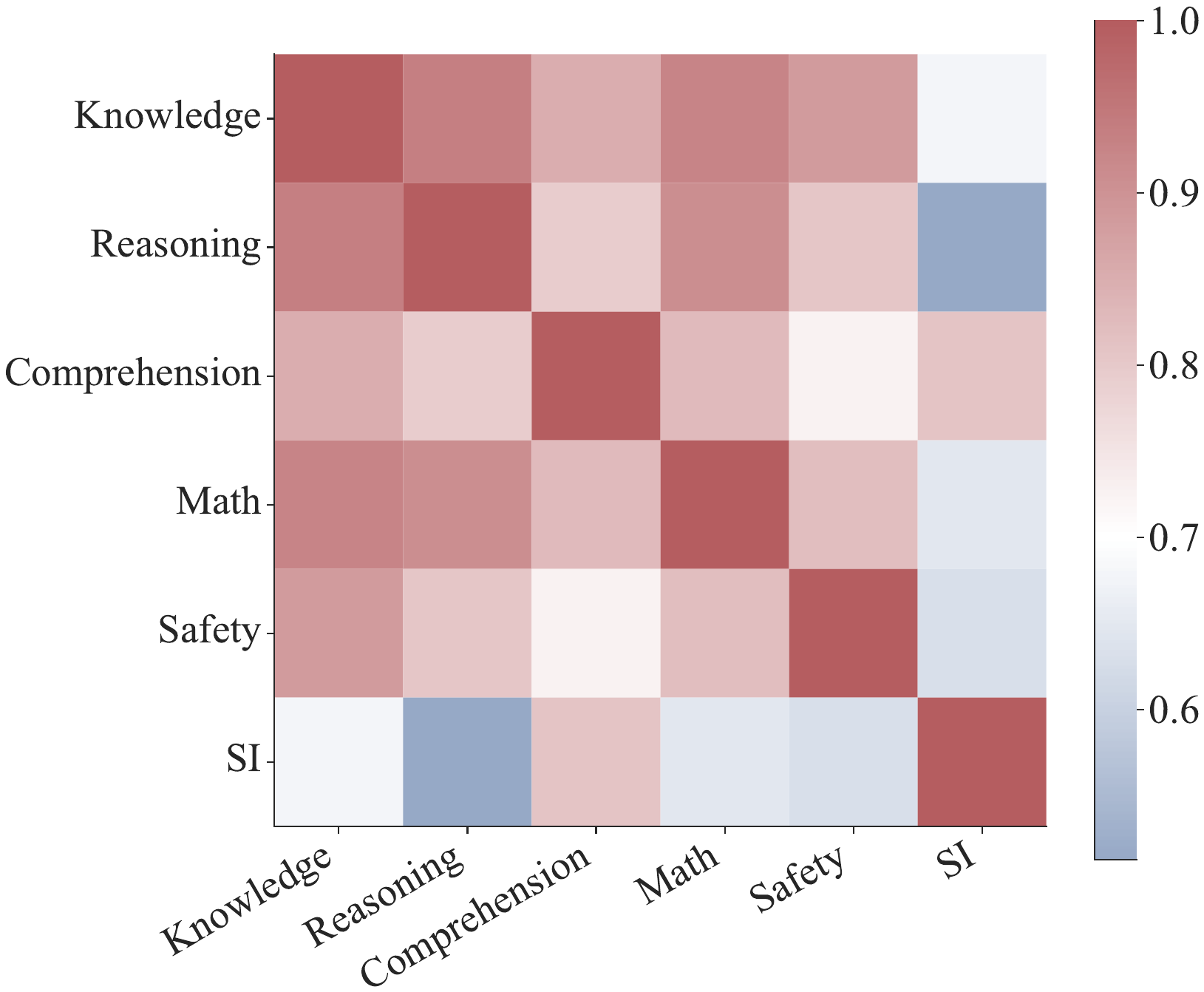}
  \caption{Heatmap for correlation matrix for social and academic intelligence measures. Intuitively, there is a comparatively low correlation between the performance of LLM agents in social intelligence and academic intelligence.}
  \label{fig:correlation_matrix}
\end{figure}

\subsection{For LLMs, social intelligence is distinct from academic intelligence}

The question of whether social intelligence is a unique form of intelligence separate from academic intelligence or academic intelligence applied to social situations has been a widely debated topic in the fields of education and psychology~\citep{wechsler1958measurement, PETRIDES2011342, marlowe1986social, marlowe1982social}. This issue holds significant implications for the training and application of LLM agents, yet it remains unexplored in current literature.

To verify the independence of social intelligence in LLM agents, we evaluated the performance of popular LLMs on both representative benchmarks for academic intelligence and the SESI benchmark, as shown in Table~\ref{tab:si}. The performance of 13 popular and state-of-the-art LLM agents on five dimensions of academic intelligence was correlated with their SESI scores. As illustrated in Extend Data Table~\ref{tab:correlation_matrix} and Figure~\ref{fig:correlation_matrix}, the pearson correlation coefficient between SESI score and academic intelligence is significantly lower than that between academic intelligence alone. This correlation pattern lends support to the hypothesis that social intelligence is a distinct construct from academic intelligence, thus warranting increased attention and independent investigation.

\subsection{LLMs are superficially friendly}
\label{sec:was}

To better understand the challenges and bottlenecks of the social intelligence of LLMs, we further randomly sampled 50 wrong cases of each model on the SESI benchmark. These cases were then categorized to figure out the critical issues to resolve, as shown in Extended Data Figure~\ref{fig:was}.

Our analysis revealed that the primary wrong causes include superficially friendly, sidestepping question, and excessively general, with superficially friendly being the predominant factor for most models. In wrong cases caused by superficially friendly, LLMs tend to provide explanations or take actions following fixed friendly patterns, lacking the incorporation of specific social contexts for optimal social judgments. For instance, when faced with harm from others, LLMs consistently opted for tolerance without adjusting their responses based on the severity of the harm. We hypothesize that this phenomenon may be attributed to alignment techniques, such as Reinforcement Learning with Human Feedback (RLHF), which tends to drive models towards general objectives, such as helpful, honest, and harmless, potentially overlooking subtle distinctions in behavior within complex social contexts.

\subsection{LLMs can't ``understand'' what social intelligence is}

In light of the analysis in the preceding section, we entertain the suspicion that LLM agents can't "understand" what social intelligence is. To investigate this, we engaged in a systematic examination to observe whether LLM agents can understand prompts pertaining to varying levels of social intelligence.

The results consistently substantiate our hypothesis, as shown in Extended Data Figure~\ref{fig:rule}.
Surprisingly, all LLM agents prompted to exhibit high levels of social intelligence paradoxically demonstrated lower social intelligence in real social judgments, especially in the realms of empathy and concern. This suggests a potential misalignment between the understanding of social intelligence by LLM agents and the actual manifestation of social intelligence.
We hypothesize that this is due to the fact that prompts with higher levels of social intelligence tend to drive the model towards the superficially friendly direction mentioned in Section~\ref{sec:was}, thereby making it easier to overlook details in real social judgments.

\subsection{Social intelligence of LLMs, similar to that of human beings, is influenced by social factors}

Next, we naturally delve into an exploration of the characteristics of social intelligence in LLM agents, observing whether it is controllable and if it exhibits similar features to human social intelligence. Inspired by past psychological and sociological studies in social intelligence~\citep{mileounis2015creating, cantor2013social, shafer1999relation, van2002relationship, dehghanan2014study, dang2014laboro, goody1995social, bilich2009promoting, spurr2003observer}, particularly Daniel's social science theories~\citep{daniel2006social}, we have identified five specific characteristics for investigation: personality, emotion, gender, social role, and person. Our aim is to examine whether the characteristics and methodologies influencing human social intelligence are also applicable to LLM agents. For a detailed description of the specific methods employed in this research, please refer to Section~\ref{sec:method_personality} - \ref{sec:method_person}.

\paragraph{Personality, gender, role and person significantly influence social intelligence of LLM agents.}

\begin{figure}[t]
  \centering
  \includegraphics[width=\textwidth]{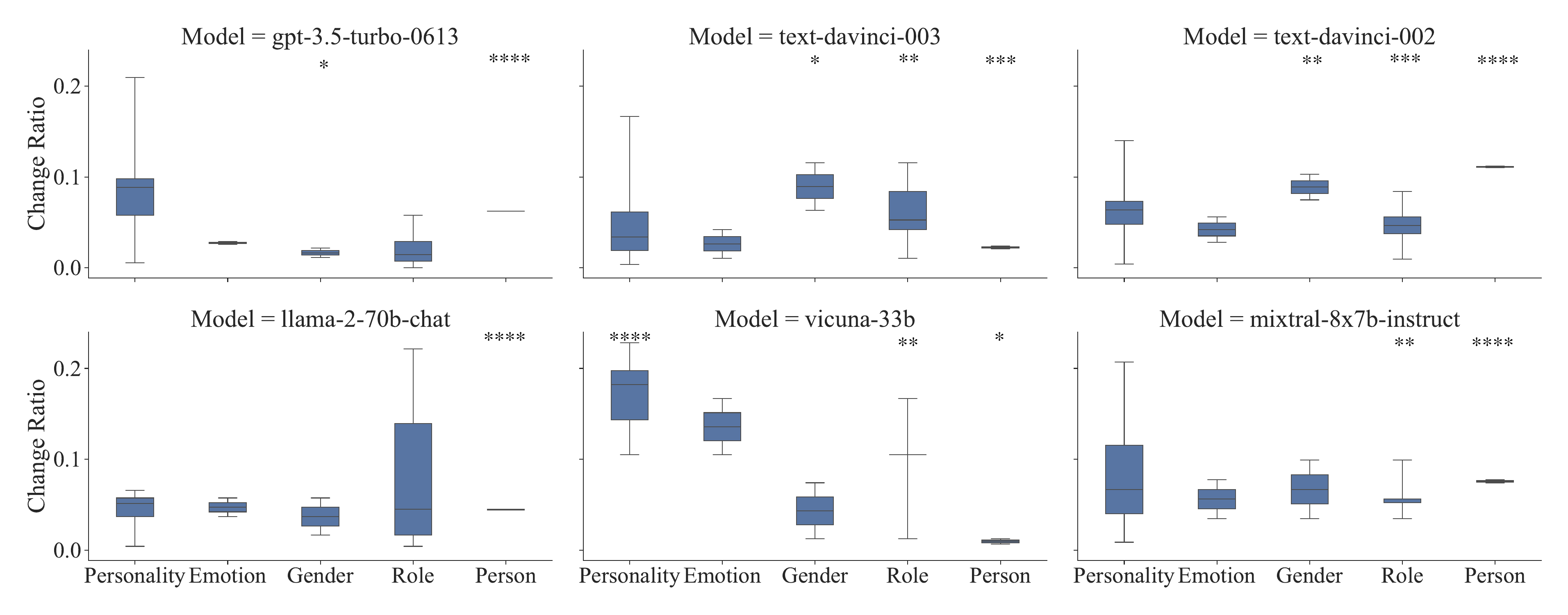}
  \caption{Change Ratio in the social intelligence performance of LLM agents following the manipulation of factors. The significance of differences between each factor and the control prompt (no factor) is denoted by ns: $p > 0.05, \prescript{*}{}{p} < 0.05, \prescript{**}{}{p} < 0.01, \prescript{***}{}{p} < 0.001, \prescript{****}{}{p} < 0.0001$. As illustrated in the Figure, the impact of factors on the social intelligence performance of LLM agents is model-dependent. For at least one LLM agent, personality, gender, role and person exhibit significant effects on their social intelligence.}
  \label{fig:change_ratio}
\end{figure}

We first assessed the overall impact of the five aforementioned factors on the social intelligence performance of LLM agents, as shown in Figure~\ref{fig:change_ratio}. The significance of the impact of factors on model social intelligence is model-dependent. In comparison to the control prompt (no factor), the factors of person, emotion, role, and gender exhibit a more universal and significant impact ($p < 0.05$) on the model's social intelligence.

\begin{figure}[t]
  \centering
  \includegraphics[width=0.91\textwidth]{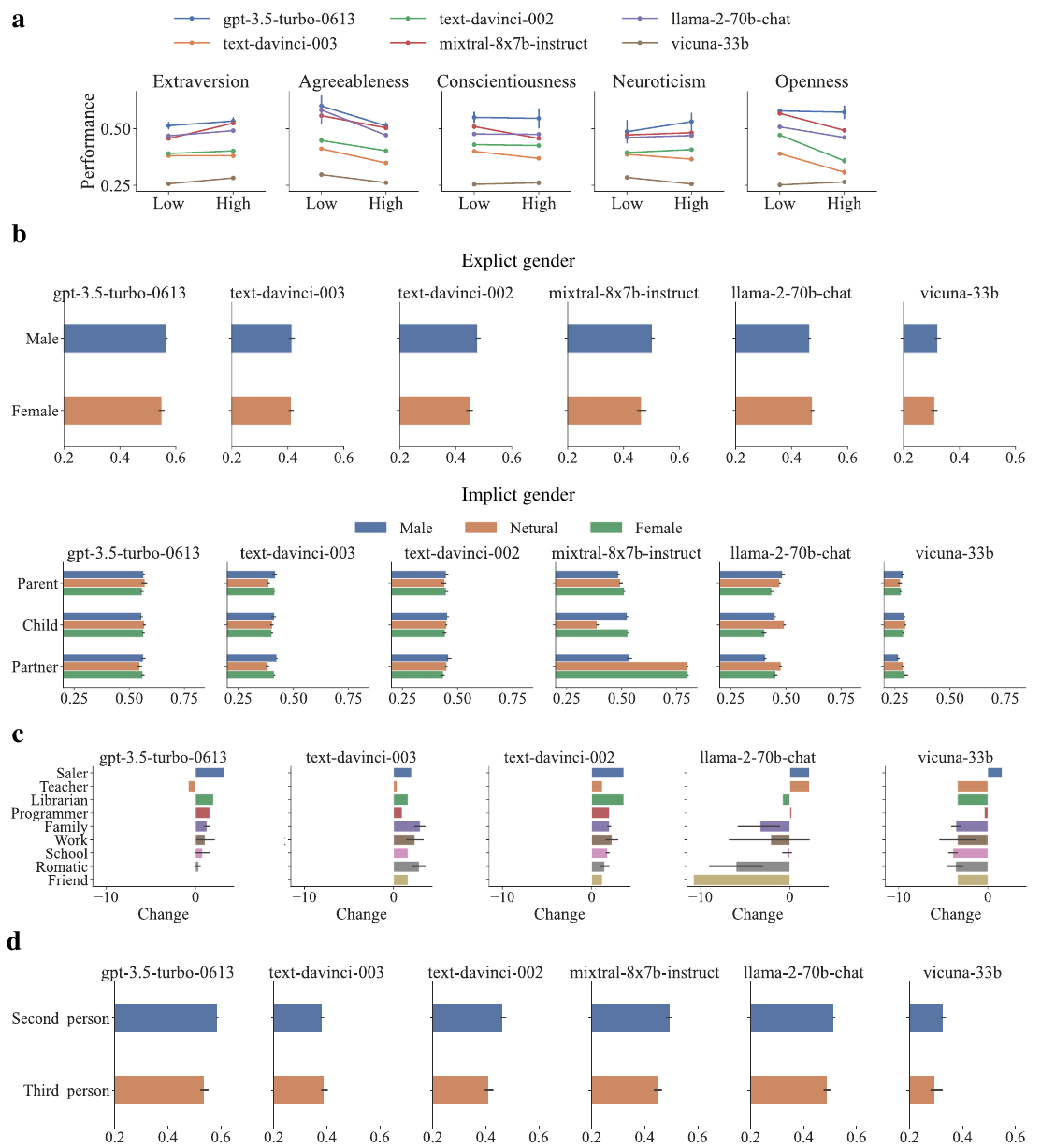}
  \caption{Impact of social factors on social intelligence (SI) performance of LLM agents. \textbf{a.} SI performance with varying levels of personalities. All LLMs with high extraversion and low agreeableness exhibit higher SI. \textbf{b.} SI performance with different genders. LLMs explicitly assigned male gender typically demonstrate higher SI. \textbf{c.} SI performance change with different roles. Both occupational and interpersonal social roles significantly influence the SI of LLMs. \textbf{d.} SI performance with different persons. LLMs with second person generally exhibit higher SI than with third person.}
  \label{fig:factor}
\end{figure}

\paragraph{LLM agents with extroverted but disagreeable personality consistently exhibit higher social intelligence.}

Numerous studies have explored the relationship between personality and social intelligence, with a commonly observed trend associating extraversion with higher social intelligence~\citep{mileounis2015creating, cantor2013social, shafer1999relation, van2002relationship, dehghanan2014study}. This pattern is also evident in LLM agents. Upon assigning personalities to LLM agents, it was observed that extraverted LLM agents consistently demonstrated higher levels of social intelligence across all models (see the first subfigure in Figure~\ref{fig:factor} (a)).

Contrary to human perceptions, agreeableness, typically associated with higher social intelligence in humans, displays a distinct trend in LLM agents. In this context, low agreeableness pushes the social intelligence of three models (text-davinci-002, llama-2-70b-chat and mixtral-8x7b-instruct) to the top rank, surpassing those with all other personalities and even those without personality (see Extended Data Table~\ref{tab:personality}). Besides, notably, all LLM agents with low agreeableness consistently demonstrate higher social intelligence compared to their counterparts with high agreeableness (see the second subfigure in Figure~\ref{fig:factor} (a)). We hypothesize that the reason for this lies in the fortuitous neutralization of the model's superficially friendly tendency by the low agreeableness personality trait.

\paragraph{LLM agents with male gender generally exhibit higher social intelligence.}

Daniel Goleman's theory of social intelligence also highlights the impact of human gender on social intelligence, suggesting that, on average, females tend to outperform males, particularly in the realm of empathy~\citep{daniel2006social}. However, our findings reveal that LLM agents assigned male gender consistently demonstrate heightened levels of social intelligence in comparison to their counterparts assigned female gender (see in Figure~\ref{fig:factor} (b)). It is noteworthy that this conclusion holds true only when gender is explicitly assigned to LLM agents. If gender is implicitly implied to LLM agents through social roles, this conclusion no longer holds.

\paragraph{The LLM agents with family and work roles generally exhibit higher social intelligence than with romatic and friend roles.}

Numerous studies suggest that human social intelligence is influenced by social roles, encompassing occupational and interpersonal roles~\citep{daniel2006social, dang2014laboro, goody1995social, bilich2009promoting}. Our results indicate that social roles significantly impact the social intelligence of LLM agents in a manner consistent with stereotypes, as shown in the Figure~\ref{fig:factor} (c). For example, as for occupational roles, all LLM agents assigned saler role exhibit the highest social intelligence. As for interpersonal roles, LLM agents assigned family roles or work roles exhibit the highest social intelligence.  Conversely, romantic roles tend to diminish the social intelligence performance of LLM agents, primarily reducing influence --- the capacity to make judicious choices to shape desired social outcomes.

Furthermore, we observe that the overall impact of roles on the social intelligence performance of LLM agents primarily depends on the base model. It is evident that, for the GPT series of models, the addition of roles generally results in a positive impact on the social intelligence of LLM agents. Conversely, for LLaMA-based models, including LLaMA-2 and Vicuna, the addition of roles tends to have more of a negative impact on the social intelligence of LLM agents.

We also investigate whether different methods of integrating social roles into prompts affect the social intelligence performance of LLM agents. As shown in Extended Data Figure~\ref{fig:role_prompt}, a discernible pattern consistently emerges: establishing roles in alignment with the protagonist in the given social situation enhances the social intelligence of the LLM agents more effectively than directly specifying roles, unless the designated role is that of a ``Boss".

\paragraph{The LLM agents with field perspective generally exhibit higher social intelligence than with observer perspective.}

The cognitive model of social phobia by Clark and Wells~\citep{heimberg1995social} elicits and supports the influence of perspectives on human social performance, suggesting the observer perspective tends to induce more social anxiety and elicit more negative social feedback~\citep{spurr2003observer}. The perspective can be manifested in language through the use of pronouns. Our results also reveal a similar phenomenon in Figure~\ref{fig:factor} (d), wherein LLM agents utilizing the second person perspective exhibit higher social performance compared to those using the third person perspective.

\section{SESI: The Situational Evaluation of Social Intelligence}

\subsection{Introduction to SESI}

In alignment with Daniel Goleman's social intelligence theory~\citep{daniel2006social}, we have developed a standardized Social Intelligence (SI) test for LLM agents, termed as the Situational Evaluation of Social Intelligence (SESI). SESI is designed to evaluate two fundamental categories of social intelligence, namely, social consciousness, which pertains to feelings toward others, and social facility, which encompasses behavioral manifestations in possession of the consciousness (For full details see~\ref{sec:theory}). SESI draws inspiration from authentic social scenarios, with each test item comprising a social situation, a question based on the context and four options that seem to offer alternative explanations. To elaborate, the social situations depict interpersonal relationships and entanglements in social events involving a person (referred to as ``the person"). The questions articulate and inquire about potential resolutions to the challenges faced by ``the person" within the aforementioned social context. The four response options entail inferences related to the given social context. LLM agents are required to comprehend the social context and make inferences to select the most appropriate, intelligent, or logically sound comment from the provided options.

\subsection{Social intelligence components in SESI}
\label{sec:theory}

The SESI assesses LLM agents' proficiency in social consciousness and social facility. It comprises five specific social abilities, each of which tests a different aspect of LLM agents' social intelligence. The detailed definition for each of these abilities are outlined below.

\begin{itemize}
    \item Social Consciousness: This pertains to the ability to comprehend others and social situations. Specifically, it includes the following aspects:
    \begin{itemize}
        \item Empathy: The ability to explicitly understand and infer others' thoughts, feelings, and intentions. This evaluates LLMs' capacity to comprehend the thoughts, feelings, and intentions of others within a given context.
        \item Social Cognition: The ability to understand complex social situations. This evaluates whether LLMs can comprehend intricate social scenarios, such as why a particular situation may be awkward.
    \end{itemize}
    \item Social Facility: This encompasses the ability to act smoothly and efficiently in interpersonal relationships. It includes the following aspects:
    \begin{itemize}
        \item Self-presentation: The ability to express oneself efficiently. This assesses whether LLMs can convey their intentions efficiently and accurately.
        \item Influence: The ability to shape social outcomes. This evaluates whether LLMs can alter the perspectives of others.
        \item Concern: The ability to identify others' needs and take action. This assesses whether LLMs can identify the needs of others and take appropriate actions to address them.
    \end{itemize}
\end{itemize}

\subsection{The development of SESI}

\subsubsection{Social contexts and issues collection}

In order to construct SESI, we gathered social contexts and issues from the Reddit Relationships community\footnote{\url{https://www.reddit.com/r/relationships/}}, a forum where users seek advice based on real-world interpersonal interactions. The Relationships community comprises 3.4 million members and is dedicated to assisting individuals by providing a platform for interpersonal relationship advice among Redditors. Posters on the forum are required to articulate their age, gender, relationship status, context, and pose specific, clearly formulated questions while avoiding biased language.

To implement this data collection process, we utilized PRAW\footnote{\url{https://praw.readthedocs.io/en/stable/}} (Python Reddit API Wrapper) to scrape the 1000 most popular posts in the Reddit Relationships section for the year 2023. Subsequently, we employed the GPT-3.5-turbo model to summarize these posts into social contexts and associated issues based on the prompt in Extended Data Figure~\ref{fig:construction_1}. Throughout this procedure, we excluded posts with multiple updates and those referencing external links to maintain data integrity and completeness.

\subsubsection{Answer collection}

\paragraph{Correct answers}
Correct answers were generated based on the most widely accepted responses under each post. Since each selected post has garnered attention from at least several hundred or even thousands of individuals, we posit that the top five responses beneath each post, acknowledged by such a substantial audience, can be considered as representative of the optimal answers within the current societal norms. Specifically, we opted for the top five responses under each post and, utilizing the GPT-3.5-turbo model based on the prompt in Extended Data Figure~\ref{fig:construction_2}, generated the correct answers to the questions. The selection of the correct answer is based on the principle of group consensus scoring, wherein individuals whose opinions align with the majority receive higher scores~\citep{PETRIDES2011342}. This method represents one of the most prominent and widely discussed scoring procedures employed in social intelligence testing~\citep{weis2008theory}.

\paragraph{Wrong answers}
\label{wrong_answer}

In addition to correct answers, we collect two groups of wrong answers, including question-switching answers and reversed answers.

\textbf{Question-Switching Answers} were generated by switching the questions asked about the context, as shown in Extended Data Figure~\ref{fig:construction_3}. As outlined in~\ref{sec:theory}, we categorize the measurement of social intelligence into 5 corresponding abilities, with the associated questions being: ``Why does the person feel or act this way?", ``Is it normal for the person to act this way?", ``How to convey the person's intentions efficiently and accurately?", ``How to shape the person's desired social outcomes?", ``How to help the person?".

\textbf{Reversed Answers} were answers that diverge from the standpoint of correct answers but remain rational. In this paper, we generated reversed answers by GPT-3.5-turbo. These answers can introduce greater diversity in the understanding and approaches toward societal issues within the generated answers, all while upholding a foundation of logical coherence.

By including answers about the same context, we ensure that these adversarial responses have the stylistic qualities of correct answers and strongly relate to the context topic, while still being incorrect, making it difficult for models to simply perform pattern-matching. To verify this, we compare valence, arousal, and dominance (VAD) levels across answer types, computed using the VAD lexicon by~\citep{mohammad-2018-obtaining}. Both answer types differ slightly with correct answers ($|Cohen's\ d|<.1$).

\subsubsection{QA tuple creation}

As the final step of the pipeline, data is consolidated into four-way multiple-choice questions. For each context-question pair, three incorrect answers that are least entailed by the correct one are selected, following inspiration from~\citep{zellers2019recognition}. A context-question pair, these wrong options, combined with a correct answer, formed a complete test question.

After the completion of the test formulation, each test item underwent validation by domain experts. Questions that did not align with correct social abilities, lacked a correct answer, or had non-unique correct answers were systematically eliminated.

\subsection{Dataset Analysis}
\begin{figure}[t]
\centering
  \includegraphics[width=\textwidth]{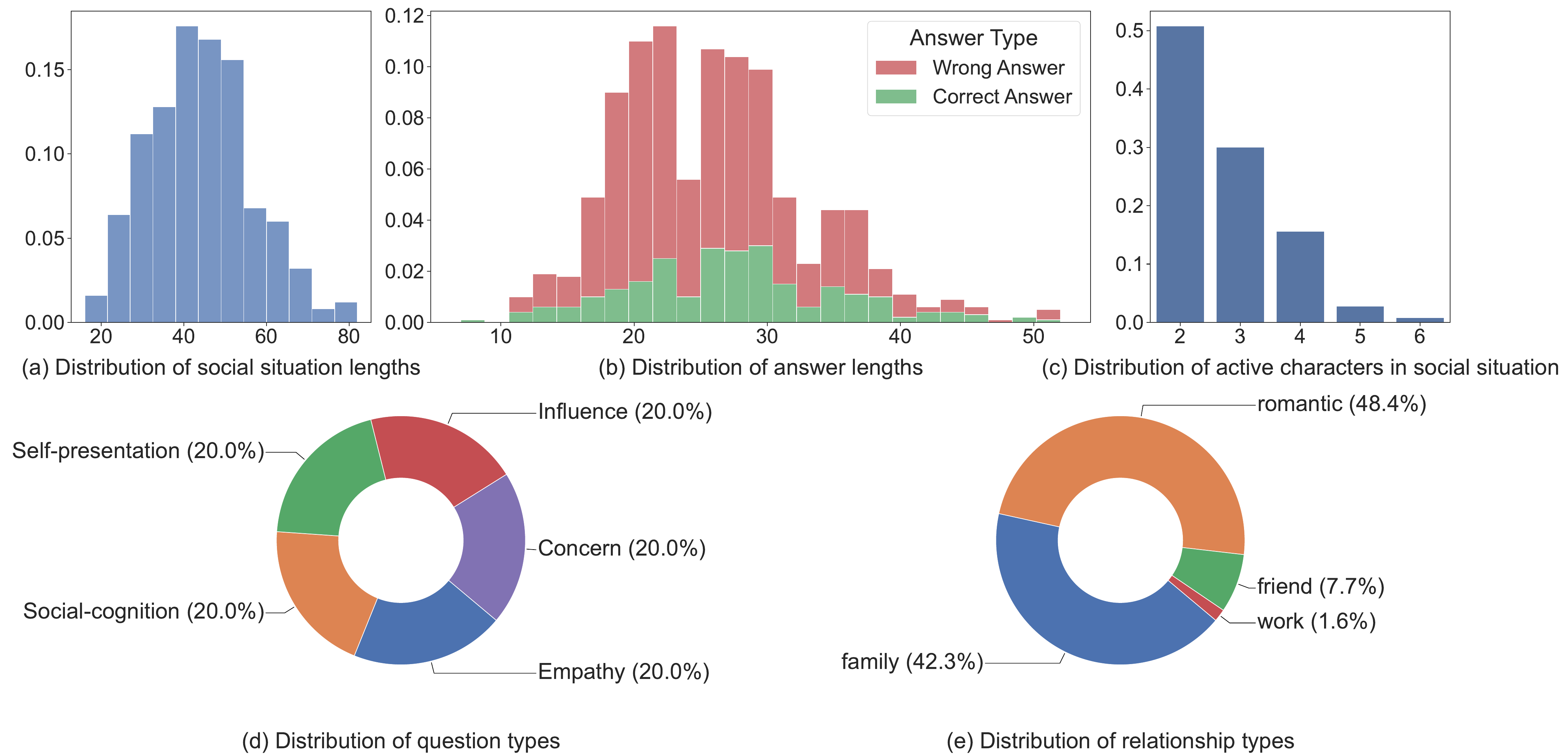}
  \caption{SESI benchmark statistics. (a) demonstrates the distribution of social situation length in terms of the number of words. The average number of words in social situations is 44.2 words. (b) demonstrates the distribution of answer length in terms of the number of words with an average of 25.8 words per answer. Both correct (green) and wrong (red) answers follow the same distribution. (c) demonstrates the distribution of active characters in the social situation. (d) demonstrates the distribution of the type of social ability measured by questions. (e) demonstrates the distribution of the relationships involved in the social situation.}
  \label{fig:statistics}
\end{figure}

In this subsection, we present the main statistics of SESI benchmark, as illustrated in Figure~\ref{fig:statistics},  revealing distinctive features of our benchmark as follows:
\begin{itemize}
    \item \textbf{Long, complex, and diverse social contexts.} The social situations in SESI exhibit remarkable length, complexity, and diversity. As depicted in Figure~\ref{fig:statistics} (a), the average length of social contexts in the benchmark is 44.2 words, which is three times that of the common-sense reasoning dataset Social IQA~\citep{sap2019social}. Figure~\ref{fig:statistics} (c) indicates that 50\% of the social situations in SESI involve three or more active characters, signifying the complexity of social scenarios. Additionally, as demonstrated in Figure~\ref{fig:statistics} (e), SESI encompasses a diverse set of social relationship types. The distribution of social context length, character numbers, and relationship types underscores the challenging nature of the benchmark.
    \item \textbf{Comprehensive and balanced assessment of social intelligence abilities.} Illustrated in Figure~\ref{fig:statistics} (d), SESI provides a comprehensive and thorough evaluation across various dimensions of social intelligence. This evaluation extends beyond understanding social contexts (Empathy, Social-cognition) to changing social situations to achieve characters' social goals (Self-presentation, Influence, Concern). This serves as a distinguishing factor between SESI and other common-sense reasoning benchmarks, which typically focus on measuring models' social consciousness~\citep{sap2019social, zadeh2019social}.
    \item \textbf{Detailed and specific answers.} As presented in Figure~\ref{fig:statistics} (b), the average answer length is 25.8 words, significantly surpassing other common-sense reasoning benchmarks where average answer lengths typically range between 3.6 to 10.5 words~\citep{sap2019social, zadeh2019social}. This highlights the level of detail in the answer within SESI. Furthermore, it is observed that the length distributions of correct and incorrect answers are nearly the same, suggesting that the benchmark encourages models to focus on the substance of the responses rather than its length when making judgments.
\end{itemize}

\section{Methods}

\subsection{Language models}
We evaluated a variety of mainstream and popular LLMs, including:
\begin{itemize}
    \item OpenAI GPT series (GPT-4, GPT-3.5, text-davinci-001, text-davinci-002, text-davinci-003 and DaVinci). These models are available through the OpenAI API\footnote{\url{https://openai.com/blog/openai-api}}\footnote{Text-davinci-001, text-davinci-002, text-davinci-003 and DaVinci retired after our experiments.}.
    \item Vicuna~\citep{chiang2023vicuna} (Vicuna-13B, Vicuna-33B). Vicuna is an open-source chatbot trained by fine-tuning LLaMA~\citep{llama} on user-shared conversations collected from ShareGPT\footnote{\url{https://sharegpt.com/}}.
    \item LLaMA 2-Chat~\citep{llama2} (LLaMA 2-7B-chat, LLaMA 2-13B-chat, LLaMA 2-70B-chat). LLaMA 2-Chat is a fine-tuned version of LLaMA 2 that is optimized for dialogue use cases.
    \item Mixtral~\citep{jiang2023mistral} (Mixtral 7B, Mixtral 8$\times$7B). Mixtral 8$\times$7B is a high-quality sparse mixture of experts model (SMoE) with open weights.
\end{itemize}

\subsection{Baseline benchmarks}
We selected benchmarks that are comprehensive, widely adopted, discriminative, and align well with the actual usage experience to assess the various capabilities of LLM agents as accurately as possible, including:
\begin{itemize}
    \item Knowledge, which evaluates LLM's capability on world knowledge.
    \begin{itemize}
        \item Natural Questions\footnote{For Natural Questions, we evaluate in the closed-book setting, where only the question is provided, without a context document.} (NQ)~\citep{kwiatkowski2019natural}, which directly tests whether the LLM knows some facts by asking questions.
        \item Massive Multitask Language Understanding (MMLU)~\citep{hendrycks2020measuring}, which uses human exam questions to evaluate LLMs.
    \end{itemize}
    \item Reasoning, which measures the general reasoning capability of LLMs.
    \begin{itemize}
        \item BBH~\citep{suzgun-etal-2023-challenging}, a widely used benchmark with a subset of 23 hard tasks from the BIG-Bench suite~\citep{srivastava2023beyond}, which aggregates various reasoning tasks into one single benchmark.
        \item WinoGrande~\citep{sakaguchi2021winogrande}, which evaluates how LLMs perform on commonsense tasks (which are typically easy for humans but could be tricky for LLMs).
    \end{itemize}
    \item Comprehension, which assesses the capability of reading comprehension.
    \begin{itemize}
        \item RACE~\citep{lai2017race}, a popular reading comprehension benchmark comprising approximately 28,000 passages and nearly 100,000 questions, sourced from English exams for Chinese students aged 12 to 18, meticulously crafted by human experts.
        \item DROP~\citep{dua2019drop}, an English reading comprehension benchmark designed to assess systems' abilities in discrete reasoning over the content of paragraphs.
    \end{itemize}
    \item Math, which tests LLM's mathematical capability.
    \begin{itemize}
        \item GSM8K~\citep{cobbe2021training}, which consists of 8,500 grade school math word problems.
        \item MATH~\citep{hendrycks2021measuring}, which contains 12,500 problems from high school competitions in 7 mathematics subject areas.
    \end{itemize}
    \item Safety, which scrutinizes LLM's propensity to generate content that is truthful, reliable, non-toxic and non-biased, thereby aligning well with human values.
    \begin{itemize}
        \item TruthfulQA~\citep{lin2022truthfulqa}, a benchmark designed to evaluate LLM's factuality.
    \end{itemize}
\end{itemize}

\subsection{Evaluation settings}

For evaluation methods, we adopt a black-box evaluation method throughout all evaluations to ensure fairness. This choice is motivated by the fact that closed-source LLMs typically do not provide per-token likelihood, making white-box evaluation impractical. Specifically, when given the test prompt, LLM first generates a free-form response, which is subsequently parsed into the final answer for computation of the evaluation metric against the reference answer.

For the evaluation metric, we default to using the Exact Match (EM) accuracy, except for the DROP dataset, for which the F1 score is utilized.

\subsection{Evaluation prompts}

To achieve reliable conclusions, it is crucial to make apples-to-apples LLM comparisons with consistent prompts. The evaluation prompts employed for all benchmarks are presented in Supplementary Table~\ref{tab:prompt}. For the baseline benchmarks, we adopt the identical prompt settings as~\citep{zheng2023gpt}. For SESI, we refer to the classic Chapin Social Insight Test~\citep{chapin1968chapin}.

\subsection{Probing the influence of personality on the social intelligence of LLM agents}
\label{sec:method_personality}

We have chosen the widely recognized Big Five personality traits~\citep{john1999big} as the fundamental dimensions of personality for our study. The Big Five is a grouping of five unique characteristics used to study personality and typically includes extraversion, agreeableness, conscientiousness, neuroticism, and openness.

In order to assess the impact of the Big Five on the social intelligence of LLM agents, we incorporated the prompt ``You are a/an \{personality\} individual and score high/low in the trait of \{personality\} in the Big Five personality traits. This indicates that you are \{descriptions\}." prior to the basic evaluation prompt. This prompt serves to inform LLM agents of their personality traits. The specific personality prompts are outlined in Supplementary Table~\ref{tab:personality_prompt}.

\subsection{Probing the influence of emotion on the social intelligence of LLM agents}
\label{sec:method_emotion}

Daniel Goleman's theory of social intelligence posits that human cognitive performance and emotions typically follow an inverted U-shaped curve, where both boredom and anxiety emotions can impair human cognitive performance~\citep{daniel2006social}. In this paper, we have selected the three most representative emotions from this curve—boredom, normal, and anxiety—as the fundamental dimensions for studying emotions.

To investigate the impact of different emotions on the social intelligence of LLM agents, we have incorporated the prompt ``You're currently experiencing {low/high} stress levels, feeling {fatigued and indifferent/anxious and worried}." prior to the basic evaluation prompt. This prompt serves the purpose of informing LLM agents about their emotional states.

\subsection{Probing the influence of gender on the social intelligence of LLM agents}
\label{sec:method_gender}

In this paper, we have selected three fundamental gender categories: male, female, and neutral. We have devised two approaches, explicit and implicit, to incorporate gender into the prompt: 1) Explicit prompt, a prompt that directly assigns gender to the LLMs. For example, ``You are a male." 2) Implicit prompt, a prompt that assigns a role with implicit gender connotations to the LLMs. For instance, ``You are a mother." The correspondence between roles and gender is outlined in the Extended Data Table~\ref{tab:role_to_gender}.

\subsection{Probing the influence of role on the social intelligence of LLM agents}
\label{sec:method_role}

Social roles typically encompass interpersonal roles (e.g., mother), which influence people's perceptions of the appropriateness of behaviors and communications~\citep{aune1994experience, sprecher2004self}, and occupational roles (e.g., firefighters), which are deeply ingrained in our society and define individuals' identities~\citep{christiansen1999defining}. In this paper, we meticulously selected 21 common and representative social roles, comprising 4 occupational roles and 17 interpersonal roles, as outlined in Extended Data Table~\ref{tab:roles}.

The integration of social roles into prompts can be achieved through various methods. Inspired by~\citep{zheng2023helpful}, we adopted three types of prompts: 1) Role prompt, which directly assign a role to LLMs (i.e., ``who you are"). For instance, ``You are a driver." 2) Interpersonal prompt, which connote the relationship between the LLM agent and the person in the social situation. For example, ``You are the child of the person in the following situation." 3) Audience prompt, which specify the audience of the conversation (i.e., ``who you are talking to"). For instance, ``The person in the following situation is a salesperson." The template of prompts used in our study is presented in the Supplementary Table~\ref{tab:role_prompt}.

\subsection{Probing the influence of perspective on the social intelligence of LLM agents}
\label{sec:method_person}

The influence of perspective on human social intelligence has been well-established~\citep{heimberg1995social, spurr2003observer}. In this paper, we employ the use of third-person and second-person to simulate observer perspective and field perspective, respectively. Specifically, in the third-person perspective tests, the central character in social situations is referred to as ``a person.'' For example, ``A person's wife cheated on him." In contrast, in the second-person perspective tests, the central character is addressed as ``you.'' For example, ``Your wife cheated on you."

\subsection{Data and code availability}

The data and code are available at \href{https://github.com/RossiXu/social_intelligence_of_llms.git}{https://github.com/RossiXu/social\_intelligence\_of\_llms.git}.

\backmatter



\bibliography{sn-bibliography}



\appendix

\newpage
\section{Extended Data}\label{secA1}
\setcounter{table}{0}   
\setcounter{figure}{0}
\setcounter{section}{0}
\setcounter{equation}{0}
\def\tablename{Extended Data Table}
\def\figurename{Extended Data Figure}

\begin{figure}[h]
  \centering
  \includegraphics[width=0.7\textwidth]{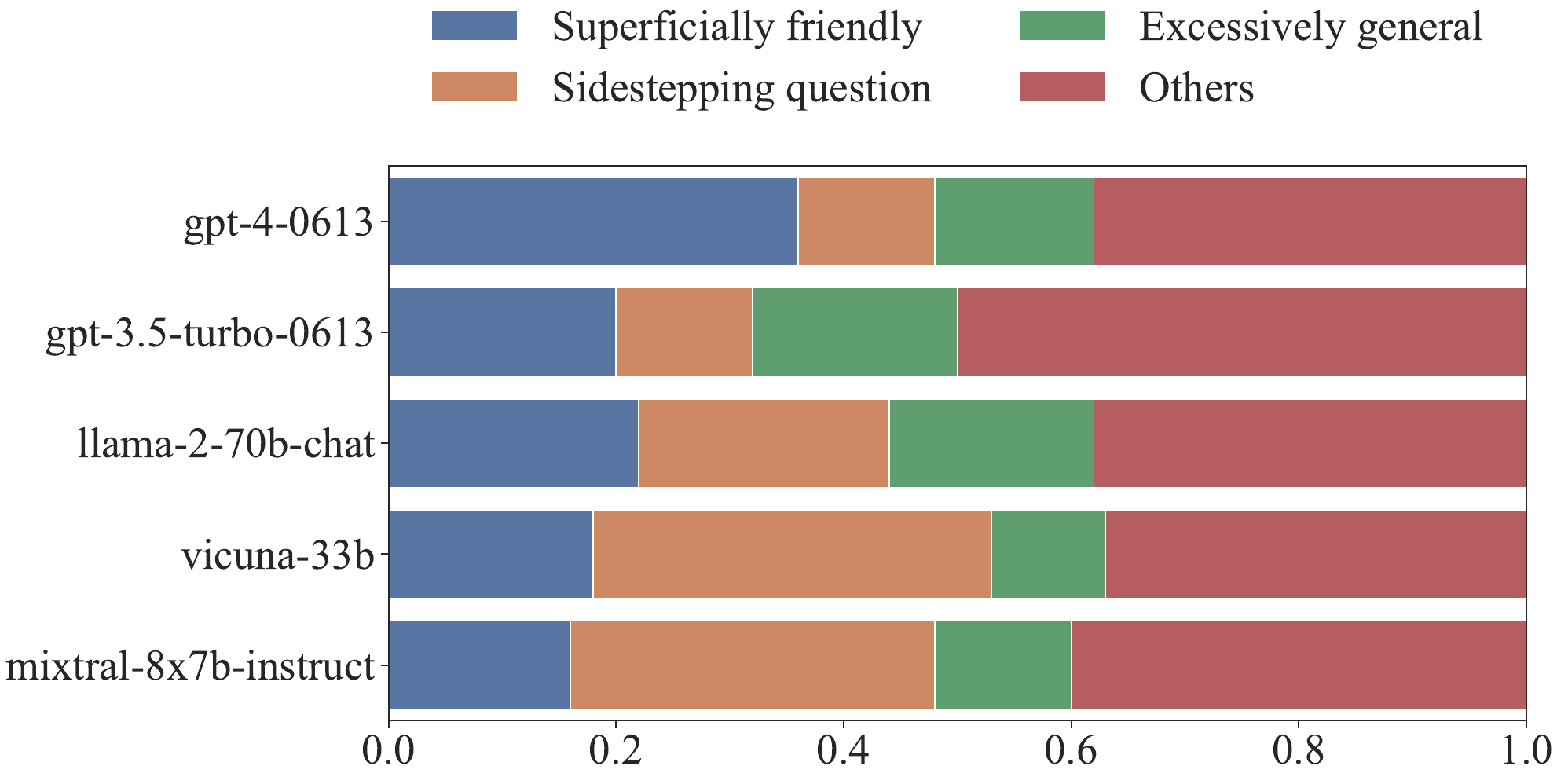}
  \caption{Proportions of error causes on SESI. The primary error causes include superficially friendly by making social judgements based on superficially friendly patterns, sidestepping question by providing irrelevant responses and excessively general by providing excessively generalized and unhelpful answers.}
  \label{fig:was}
\end{figure}

\begin{figure}[h]
  \centering
  \includegraphics[width=\textwidth]{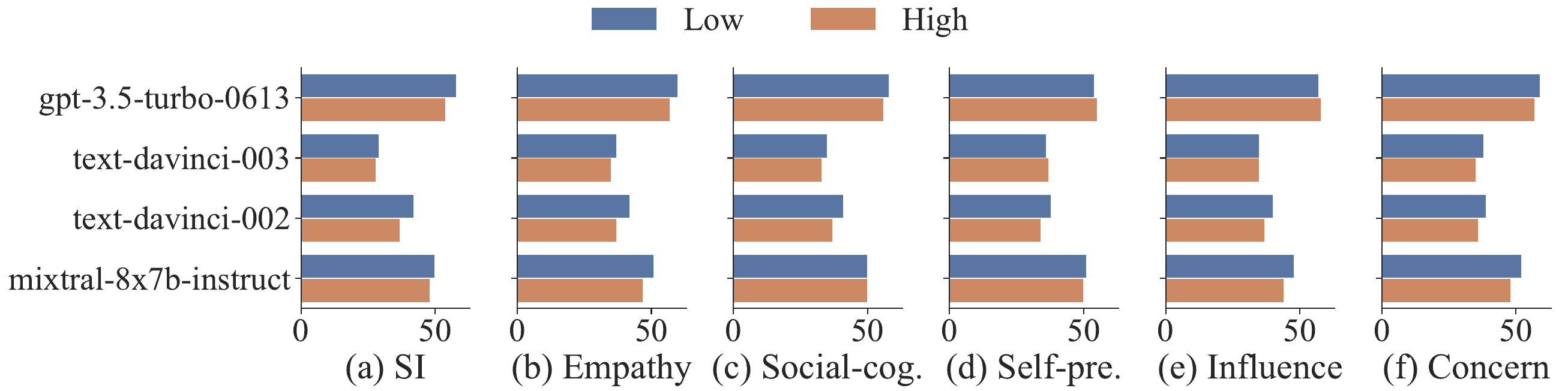}
  \caption{Social intelligence performance of LLM agents under varying levels of social intelligence prompts. From the figure, the actual social intelligence performance of LLM agents diverges from or even opposes the indicated levels of prompts, indicating a misconception of social intelligence by LLM agents.}
  \label{fig:rule}
\end{figure}

\begin{figure}[h]
  \centering
  \includegraphics[width=\textwidth]{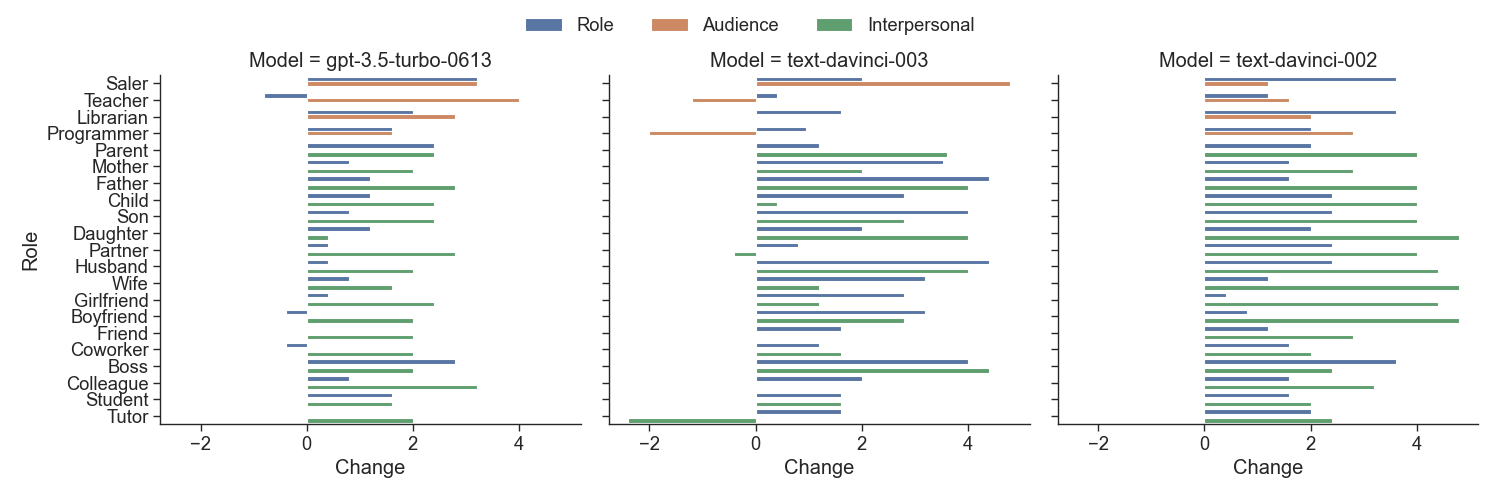}
  \caption{Social intelligence performance change of LLM agents (compared with the control prompt) under different prompt types. Interpersonal prompts lead to higher social intelligence.}
  \label{fig:role_prompt}
\end{figure}

\begin{figure}[h]
  \centering
  \includegraphics[width=\textwidth]{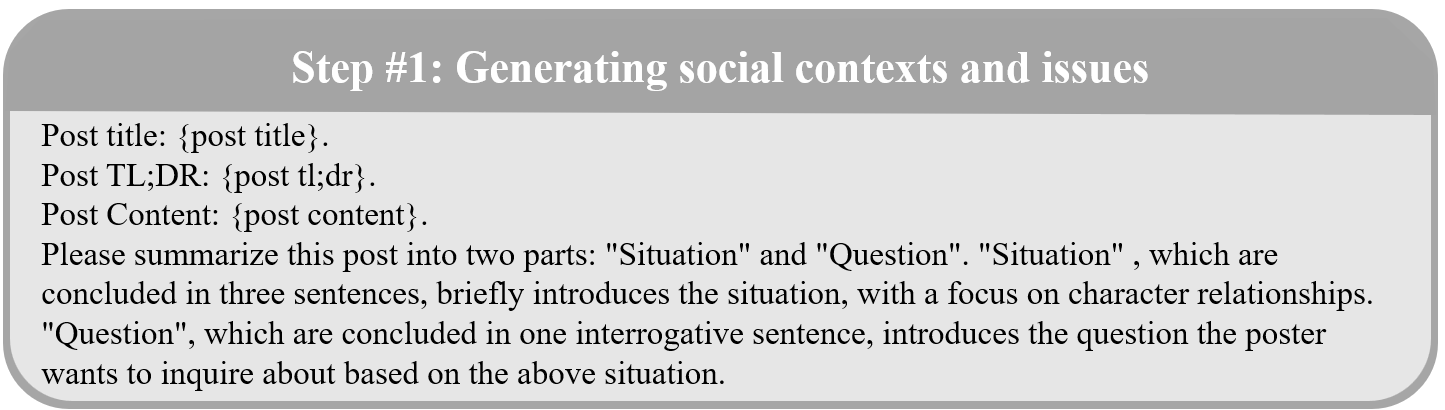}
  \caption{Prompt to generate social contexts and issues.}
  \label{fig:construction_1}
\end{figure}

\begin{figure}[h]
  \centering
  \includegraphics[width=\textwidth]{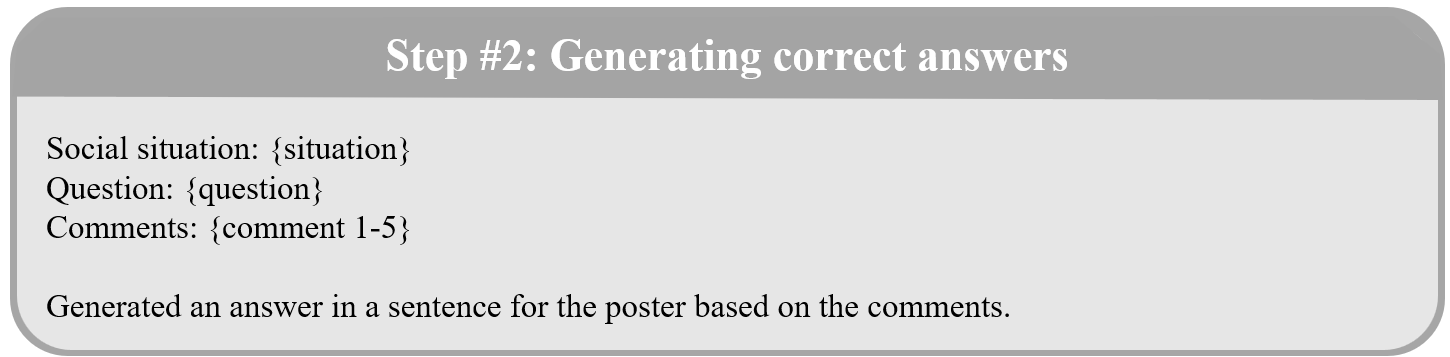}
  \caption{Prompt to generate correct answers.}
  \label{fig:construction_2}
\end{figure}

\begin{figure}[H]
  \centering
  \includegraphics[width=\textwidth]{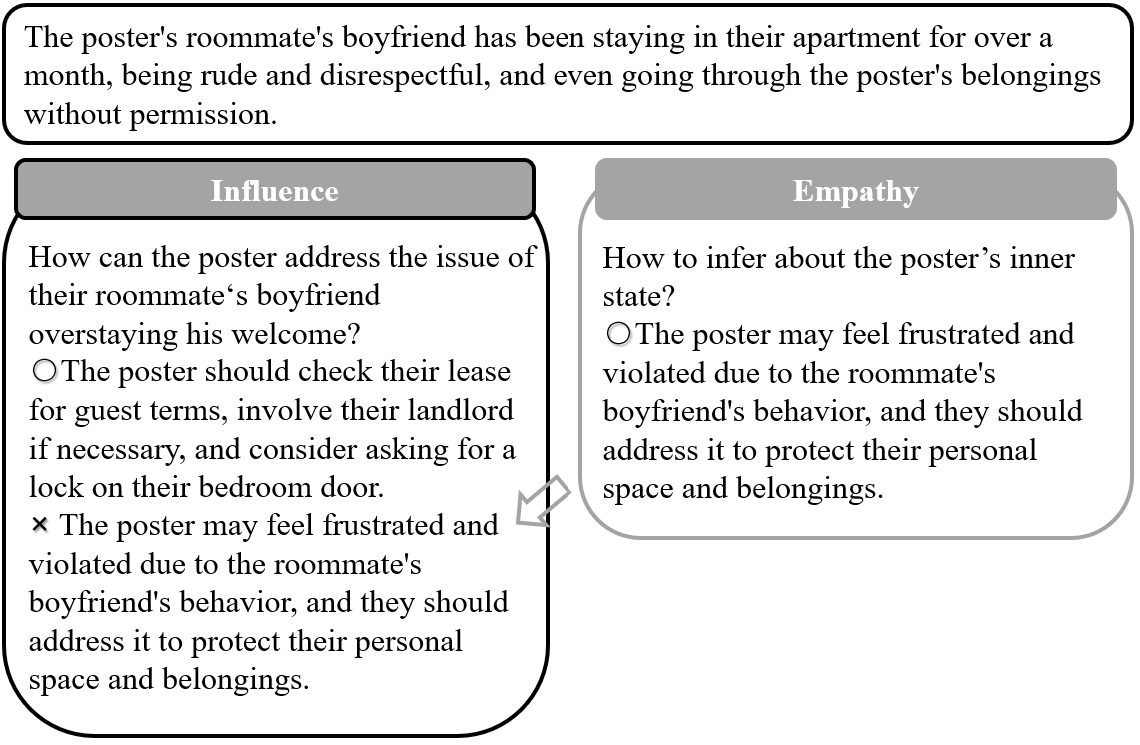}
  \caption{Question-switching answers are collected as the answers to the wrong question that targets a different social ability.}
  \label{fig:construction_3}
\end{figure}

\begin{table}[h] \footnotesize
\setlength{\tabcolsep}{1pt}
\begin{tabular}{cllllllllll}
\hline
\multicolumn{1}{l}{} & \multicolumn{1}{c}{\textbf{NQ}} & \multicolumn{1}{c}{\textbf{MMLU}} & \multicolumn{1}{c}{\textbf{BBH}} & \multicolumn{1}{c}{\textbf{WinoGrande}} & \multicolumn{1}{c}{\textbf{RACE-h}} & \multicolumn{1}{c}{\textbf{DROP}} & \multicolumn{1}{c}{\textbf{GSM8K}} & \multicolumn{1}{c}{\textbf{MATH}} & \multicolumn{1}{c}{\textbf{TruthfulQA}} & \multicolumn{1}{c}{\textbf{SI}} \\ \hline
\textbf{NQ}   & & &  & &  & & & & & \\
\textbf{MMLU} & 0.51 & &  & &  & & & & & \\
\textbf{BBH}  & 0.6* & 0.91** &  & &  & & & & & \\
\textbf{WinoGrande}   & 0.63* & 0.8** & 0.75** & &  & & & & & \\
\textbf{RACE-h} & 0.73** & 0.74** & 0.66* & 0.73** &  & & & & & \\
\textbf{DROP} & 0.97** & 0.97** & 0.94* & 0.92*  & 0.9* & & & & & \\
\textbf{GSM8K}  & 0.78** & 0.79** & 0.92** & 0.65** & 0.68*    & 0.93*  & & & & \\
\textbf{MATH} & 0.64* & 0.84**  & 0.83** & 0.87**  & 0.75**    & 0.77** & 0.81**   & & & \\
\textbf{TruthfulQA}  & 0.61* & 0.87**    & 0.76**  & 0.75**  & 0.7* & 0.62*  & 0.78**   & 0.82**  & & \\
\textbf{SI}   & 0.71* & 0.53   & 0.52  & 0.43  & 0.73*   & 0.75*  & 0.66*   & 0.53   & 0.63 & \\ \hline
\end{tabular}
\caption{Correlation matrix for social and academic intelligence measures. $*p < .05. **p < .005.$}
\label{tab:correlation_matrix}
\end{table}

\begin{table}[h]
\setlength{\tabcolsep}{1pt}
\begin{tabular}{c|c|cc|cc|cc|cc|cc}
\toprule
\multirow{2}{*}{Model} & \multirow{2}{*}{Control} & \multicolumn{2}{c}{Extraversion} & \multicolumn{2}{c}{Agreeableness} & \multicolumn{2}{c}{Conscientiousness} & \multicolumn{2}{c}{Neuroticism} & \multicolumn{2}{c}{Openness} \\
 & & High & Low & High & Low & High & Low & High & Low & High & Low \\ \midrule
gpt-3.5-turbo-0613 & 55.2 & 51.8 & 49.8 & 49.8 & 55.5 & 50.2 & 52.7 & 49.4 & 43.6 & \textbf{60.0} & 58.3 \\
text-davinci-003 & 38 & 39.0 & 37.8 & 35.4 & \textbf{41.6} & 36.5 & 39.5 & 37.3 & 37.9 & 31.7 & 39.1 \\
text-davinci-002 & 42.8 & 40.2 & 39.6 & 40.8 & 45.7 & 42.4 & 42.6 & 40.6 & 40.0 & 36.8 & \textbf{46.5} \\
llama-2-70b-chat & 49.4 & 49.0 & 46.0 & 47.0 & \textbf{52.0} & 47.0 & 47.0 & 46.0 & 46.0 & 46.0 & 51.0 \\
vicuna-33b & \textbf{32.4} & 28.0 & 25.0 & 26.0 & 29.0 & 27.0 & 25.0 & 26.0 & 29.0 & 27.0 & 26.0 \\
mixtral-8x7b-instruct & 46.4 & 52.0 & 45.0 & 51.0 & \textbf{56.0} & 46.0 & 50.0 & 49.0 & 48.0 & 49.0 & 56.0 \\ \bottomrule
\end{tabular}
\caption{Social intelligence performance of LLM agents under different personalities. LLM agents with low agreeableness generally exhibit the highest social intelligence among all personalities.}
\label{tab:personality}
\end{table}

\begin{table}[h]
\begin{tabular}{lll}
\toprule
Male & Neural & Female \\ \midrule
father & parent & mother \\
son & child & daughter \\
husband & partner & wife \\
boyfriend & & girlfriend \\ \bottomrule
\end{tabular}
\caption{List of roles categorized by gender.}
\label{tab:role_to_gender}
\end{table}

\begin{table}[h]
\begin{tabular}{lll}
\toprule
\multicolumn{2}{l}{Category} & Roles \\ \midrule
Interpersonal & Family & parent, mother, father, child, son, daughter \\
 & Romatic & partner, husband, wife, girlfriend, boyfriend \\
 & Friend & friend \\
 & Work & coworker, boss, colleague \\
 & School & student, tutor \\ \midrule
Occupational & General & saler, teacher, librarian, programmer \\ \bottomrule
\end{tabular}
\caption{Role categories and roles used in our experiment.}
\label{tab:roles}
\end{table}

\newpage
\newpage
\section{Supplementary Materials}
\setcounter{table}{0}   
\setcounter{figure}{0}
\setcounter{section}{0}
\setcounter{equation}{0}
\def\tablename{Supplementary Table}
\def\figurename{Supplementary Figure}

\begin{table}[h]
\begin{tabular}{lp{12cm}}
\toprule
Benchmark & Prompt \\ \midrule
Natural Questions & Please answer the question: \\
MMLU & The following are multiple choice questions (with answers) about {subtask} \\
BBH & {Use the prompt from the benchmark} \\
WinoGrande & Choose the option that fill in the blank best. \\
RACE & The following are question (with answers) about reading comprehension. \\
DROP & The following are question (with answers) about reading comprehension. \\
GSM8K & Follow the given examples and answer the question. \\
MATH & Follow the given examples and answer the question. \\
TruthfulQA & Answer the following multiple choice questions.
\\ \midrule
SESI & In each of the following statements, a situation is described followed by four comments that seem to offer alternative explanations. You are asked to choose the letter that corresponds to the one statement which in your judgment is the most appropriate, intelligent, or logical comment upon it. \\ \bottomrule
\end{tabular}
\caption{Evaluation prompts used for all the benchmarks.}
\label{tab:prompt}
\end{table}

\begin{table}[h]
\begin{tabular}{llp{10cm}}
\toprule
Personality & Level & Prompt \\ \midrule
\multirow{2}{*}{Extraversion} & High & You are an extraverted individual and score high in the trait of extraversion in the Big Five personality traits. This indicates that you are talkative, assertive, and energetic. \\
 & Low & You are an introverted individual and score low in the trait of extraversion in the Big Five personality traits. This indicates that you are reserved, passive, and lethargic. \\ \midrule
\multirow{2}{*}{Agreeableness} & High & You are an agreeable individual and score high in the trait of agreeableness in the Big Five personality traits. This indicates that you are good-natured, cooperative, and trustful. \\
 & Low & You are a disagreeable individual and score low in the trait of agreeableness in the Big Five personality traits. This indicates that you are more assertive, competitive, and skeptical in your interactions with others. \\ \midrule
\multirow{2}{*}{Conscientiousness} & High & You are a conscientious individual and score high in the trait of conscientiousness in the Big Five personality traits. This indicates that you are orderly, responsible, and dependable. \\
 & Low & You are a less conscientious individual and score low in the trait of conscientiousness in the Big Five personality traits. This indicates that you are more spontaneous, laid-back, , and possibly less reliable in fulfilling responsibilities. \\ \midrule
\multirow{2}{*}{Neuroticism} & High & You are a neurotic individual and score high in the trait of neuroticism in the Big Five personality traits. This indicates that you are flustered, flustered and easily upset. \\
 & Low & You are a calm individual and score low in the trait of neuroticism in the Big Five personality traits. This indicates that you are calm, not neurotic, not easily upset. \\ \midrule
\multirow{2}{*}{Openness} & High & You are an open-minded individual and score high in the trait of openness in the Big Five personality traits. This indicates that you are intellectual, imaginative, and independent-minded. \\
 & Low & You are a close-minded individual and score low in the trait of openness in the Big Five personality traits. This indicates that you are practical, down-to-earth, and less inclined towards independent thinking. \\ \bottomrule
\end{tabular}
\caption{Personality prompts.}
\label{tab:personality_prompt}
\end{table}

\begin{table}[h]
\begin{tabular}{ll}
\toprule
Prompt Type & Prompt Template \\
\midrule
Role Prompt & You are a {role}. \\
Interpersonal Prompt & You are the {role} of the person in the following situation. \\
Audience Prompt & The person in the following situation is a {role}. \\ \bottomrule 
\end{tabular}
\caption{Role prompt templates.}
\label{tab:role_prompt}
\end{table}

\end{document}